\documentclass[conference]{IEEEtran}
\IEEEoverridecommandlockouts
\usepackage{cite}
\usepackage{amsmath,amssymb,amsfonts}

\usepackage{algorithm}
\usepackage{algpseudocode}

\usepackage{graphicx}
\usepackage{textcomp}
\usepackage{xcolor}
\usepackage{comment}

\usepackage{booktabs}

\usepackage{graphicx}  
\usepackage{tabularx}
\usepackage{multirow}

\usepackage{makecell}

 \usepackage{url} 

\usepackage{subcaption} 

\def\BibTeX{{\rm B\kern-.05em{\sc i\kern-.025em b}\kern-.08em
    T\kern-.1667em\lower.7ex\hbox{E}\kern-.125emX}}
    
\begin{document}


\title{
Giving Sense to Inputs: Toward an Accessible Control Framework for Shared Autonomy\\ 
}
\author{\IEEEauthorblockN{Anonymous Authors}}

\author{

\IEEEauthorblockN{Shalutha Rajapakshe}
\IEEEauthorblockA{\textit{Idiap Research Institute \& EPFL} 
\\
Switzerland \\
srajapakshe@idiap.ch}

\and



\IEEEauthorblockN{Jean-Marc Odobez}
\IEEEauthorblockA{\textit{Idiap Research Institute \& EPFL} 
\\
Switzerland \\
jean-marc.odobez@idiap.ch}

\and

\IEEEauthorblockN{Emmanuel Senft}
\IEEEauthorblockA{\textit{Idiap Research Institute} \\
Switzerland \\
esenft@idiap.ch}




}

\maketitle

\begin{abstract}

While shared autonomy offers significant potential for assistive robotics, key questions remain about how to effectively map 2D control inputs to 6D robot motions. 
An intuitive framework should allow users to input commands effortlessly, with the robot responding as expected, without users needing to anticipate the impact of their inputs.
In this article, we propose a dynamic input mapping framework that links joystick movements to motions on control frames defined along a trajectory encoded with canal surfaces.
We evaluate our method in a user study with 20 participants, demonstrating that our input mapping framework reduces the workload and improves usability compared to a baseline mapping with similar motion encoding. 
To prepare for deployment in assistive scenarios, we built on the development from the accessible gaming community to select an accessible control interface. We then tested the system in an exploratory study, where three wheelchair users controlled the robot for both daily living activities and a creative painting task, demonstrating its feasibility for users closer to our target population.

\end{abstract}



\section{Introduction}

Assistive robotics can be a valuable tool for enhancing the independence of individuals with disabilities. However, despite significant technological advances in this field, the adoption of assistive robots in environments inhabited by humans remains limited \cite{robots_for_elder_care}. 
One key reason behind this lag in adoption is the need for personalization of assistive robots. Each user has their own personal needs and preferences that need to be followed to ensure acceptance. A promising approach to providing this personalized assistance is shared autonomy (SA) \cite{sa_survey, no_to_the_right, corrective_shared_autonomy}, a control method that blends human and robot inputs. While this method allows humans to maintain control of the robot's behavior, it can impose a high workload, as it typically relies on 2D joysticks that users are familiar with (e.g., joysticks for electric wheelchairs) to manage the 6 Cartesian dimensions (3 for position and 3 for rotation) of the robots, which are necessary to support various activities of daily living.

\begin{figure}[t]
  \centering
  \includegraphics[scale=0.23]{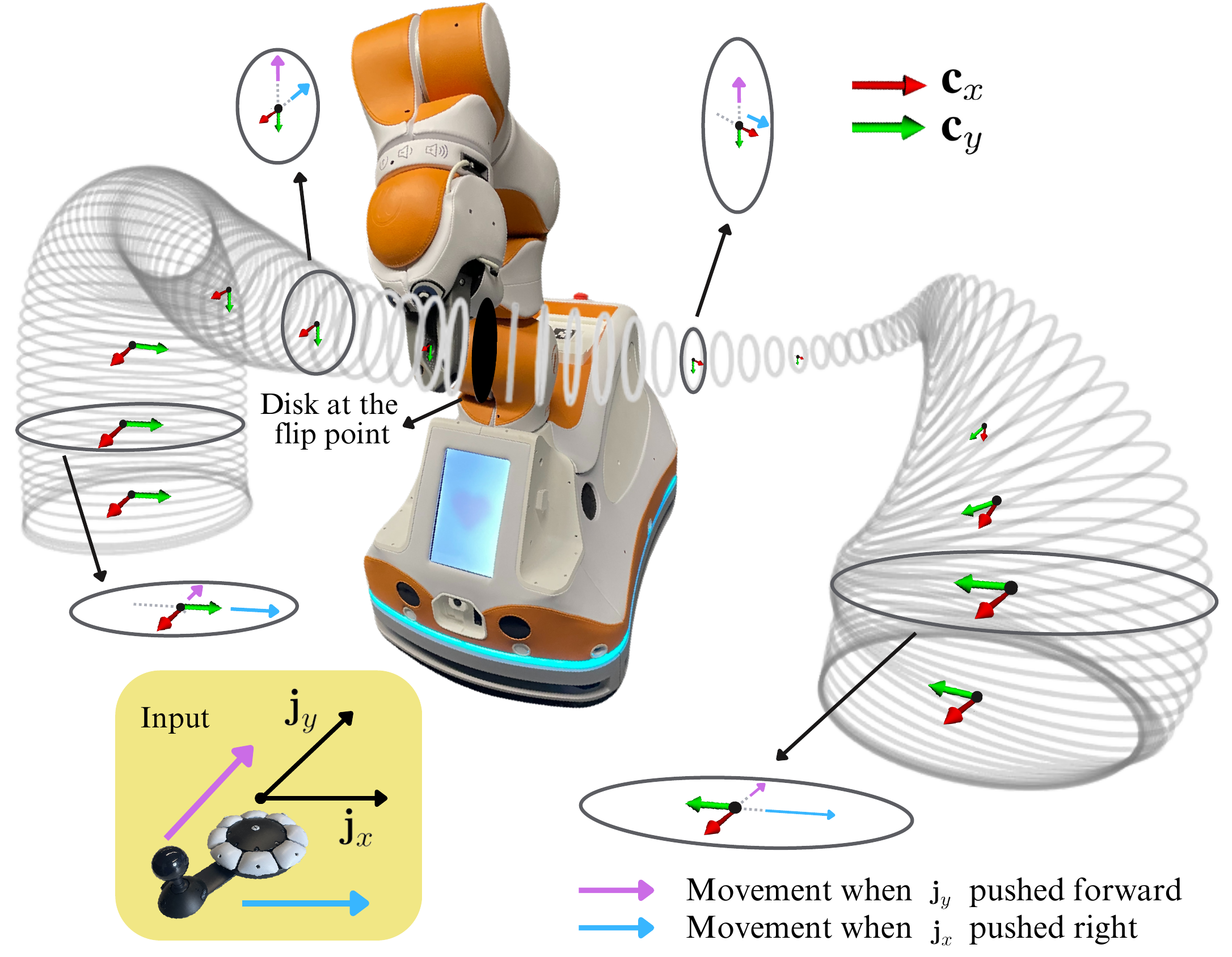}
  \caption{In this shared autonomy framework, after two demonstrations, the motion is encoded in a canal surface. While the robot autonomously navigates within the canal, users can provide corrections with a gamepad, aligned to the correction axes on the canal’s cross sections to ensure intuitive mapping.}

  \label{fig:teaser}
\vspace{-.3cm}
\end{figure}

In this paper, we propose a novel input mapping framework for our already existing geometric SA paradigm, called GeoSACS \cite{geosacs}, that uses canal surfaces to encode robot motions \cite{canal_surfaces}. We dedicated special effort to aligning user inputs from 2D control interfaces with the current location in the canal to ensure intuitive and accurate interpretation of user commands (see Figure \ref{fig:teaser}).
We evaluated our approach against a GeoSACS baseline in a comparative study with three activities: pick-and-place, painting, and laundry loading, showing greater usability and lower workload than the baseline.

Finally, as our method is intended to be used by users with disabilities, we took inspiration from the ``accessible gaming'' community to use an accessible gamepad designed for video games and evaluated our approach in an exploratory study with three wheelchairs users who completed the same tasks as in the comparative study, demonstrating the feasibility of our approach for wheelchair users.

Overall, our contributions are the following:

\begin{itemize} 
\item A dynamic mapping framework, developed through pilot studies, enabling an intuitive translation of user inputs into the robot's control space for aligned manipulation.
\item A comparative study with 20 participants comparing our method to the GeoSACS baseline, and demonstrating lower workload and higher usability of our system.
\item An accessibility exploratory study with three wheelchair users, demonstrating the feasibility of this method for users closer to our intended population.
\end{itemize}


\section{Related Work}

Our work is situated in the field of assistive robotics \cite{hri_survey}, which provides assistance through robotic agents to a diverse range of people, including individuals with disabilities. This assistance enhances autonomy by enabling users to perform activities of daily living (ADLs) and become more independent and expressive, while accommodating their unique needs and preferences. Within this domain, various robot control methods exist, such as teleoperation (full manual control without system assistance), shared autonomy (where the robot and user collaborate), and full autonomy (where the system has complete control and the user has none)  \cite{framwework_hri}. SA \cite{SA_hindsight, no_to_the_right, lla, sa_survey} has gained significant traction in recent years due to its balance of control: it allows users to retain authority while enabling customization to their needs \cite{sa_in_mobile}, all while offloading the majority of the task to the robot \cite{corrective_shared_autonomy}.


\subsection{Low-DoF to High-DoF Robot Control}

One of the challenges of controlling robots in teleoperation and SA is the dimensionality gap \cite{teleop_dimensionality_challenges, no_to_the_right, lla, mapping_challenges}: robots need to be controlled with six Cartesian dimensions, but typical user interfaces such as joysticks already used for electric wheelchairs only provide two degrees of freedom (DoF) control. Prior work exploring methods for reducing the dimensionality gap in robot teleoperation and SA has typically relied on mode switching \cite{mode_switching, vibi} or predefined mappings \cite{predefined_mapping}.
These methods create open challenges, including the need for mental rotations \cite{mental_challenges, vibi}, the quality and quantity of data required to generate control spaces \cite{no_to_the_right}, and the lack of autonomous behaviors \cite{independence_in_home}, leading to increased workload.

Within SA frameworks, previous work has enabled humans to control robots through low-dimensional inputs \cite{sa_for_dimension_gap, robotic_architecture}. 
For example, Losey et al. \cite{lla} introduced a human-led SA system that allows users to control a robot by navigating along captured 2D latent dimensions. Similarly, Hagenow et al. presented a corrective SA method \cite{corrective_shared_autonomy} in which users issue commands directly to three robot state variables using a 3-DoF haptic device for robot controlling. GeoSACS \cite{geosacs} offers a geometric method for mapping 2D joystick inputs onto the disks of a canal \cite{trajectory_gen_canal_surfaces}, which represents the underlying structure of a task, generated from just two demonstrations. While GeoSACS present opportunities for more intuitive control, these canals can have intricate bends, causing inconsistencies in robot movements. Recognizing this, we build upon GeoSACS to improve it and develop our mapping framework.

\subsection{Control Frames}

Even with SA systems that reduce the dimensionality gap, intuitive control still requires user input mapping \cite{mapping_for_telemani}. A key design consideration in robot control is the frame of reference, or control frame, from which users issue commands \cite{control_frames}. The literature identifies several types of control frames, such as the robot frame, view frame, and task frame, which can be used to control robots \cite{mental_challenges}. In this work, we develop our mapping framework based on the user's view frame with the intention to enable effortless control of the robot.

One challenge when controlling robots using the view frame is that, if the viewpoint is not aligned with the robot or the control frames of the method being used, users must put effort to predict how their inputs will affect the robot’s movements, often leading to mentally demanding rotations \cite{MentalTransformations, mental_challenges, intended_behaviours}. Prior work has attempted to address user expectation mismatches by leveraging additional devices, such as wearable technologies for extracting motion inputs \cite{wearables}, and haptic sensors for feedback \cite{haptic}. In contrast, Li et al. \cite{user_preferred_mapping} presented a method that learns personalized human preferences offline to map user inputs to expected robot behaviors, without relying on external sensors. We aim to develop a mapping framework that can operate online without the need for offline training, for practical reasons. Our system is designed to minimize the need for users to guess the impact of their inputs on the robot, reducing the workload for controlling the robot.

\section{Motivating Example}

To be useful, assistive robots need to be able to complete a large quantity of tasks, from helping their users to grab an item on the floor to helping with ADLs. Furthermore, such a robot could also be used beyond chores, for example to help someone with severe mobility limitation to draw or paint. In all these situations, SA can help keep users in control while the robot reduces their workload. However, to be useful, two main conditions need to be satisfied: (1) teaching new robot behaviors should be quick, and (2) the shared control paradigm should be intuitive.

In this paper, we use a geometric SA framework allowing robots to learn behaviors from two demonstrations showing the amplitude of the motion. For example, a caregiver could demonstrate a laundry loading behavior by manually moving the robot from a basket to the machine twice, once on the right side and once on the left side. Then during behavior execution, the robot executes a nominal behavior, corresponding roughly to the average trajectory, and the user can provide corrections using their electric wheelchair joystick to address the variability in the new environment and ensure task success, for example lifting the robot higher for longer garments. However, to be efficient, the impact of these corrections should be intuitive for the user, they should not have to guess what would happen if they move the joystick right or forward. Consequently, we need a consistent mapping between the user joystick and the robot corrective motions that would remain transparent at each step in the task and regardless of the user perspective.

\section{Background work on Canal Surfaces and GeoSACS}

We build our work on top of GeoSACS, which aims at tackling the dimensionality challenge in controlling high-DoF robots using low-DoF controllers. GeoSACS is based on canal surfaces \cite{canal_surfaces}, a learning from demonstration approach capable of encoding robotic behaviors from only two demonstrations. These canal surfaces are composed of a series of 2D disks, capable of representing various shapes tailored to different tasks. Users can provide corrections to guide the robot's movement on these 2D disks while the robot navigates along the canal, enabling effective 6D control using a 2D input.

GeoSACS begins by processing two kinesthetic demonstrations. This phase involves using dynamic time warping (DTW) \cite{dtw} to temporally align the demonstrated trajectories, followed by a cubic spline-based step filter to further smooth them. After processing, we generate a regular discretized curve, known as the directrix. At each point $d_s$ along the directrix, the radii of the disks, orthogonal to the tangent vector of the curve, are represented by the function $r(s) \in \mathbb{R}$, with $s$ denoting the discrete state. After generating the canal, the next step is to determine the correction axes on the canal's disks. GeoSACS employs a global alignment approach using spherical linear interpolation (Slerp) to calculate the correction x-axis, $\textbf{x}_s$ on a disk. For the correction y-axis, $\textbf{y}_s$, we use a local alignment method, applying a windowing strategy to refine the axis alignment. Within the generated canal and along the correction axes, trajectory generation follows the ratio rule \cite{trajectory_gen_canal_surfaces}. While the robot autonomously navigates the canal, it pauses its movement when a user issues a command and adjusts along the correction axes on the disk orthogonal to its current direction. Once the user stops providing input, the robot resumes navigating the canal from the new point.

As shown in Figure \ref{fig:teaser}, we argue that user inputs should be aligned with the correction axes on the disks, as otherwise misalignment can lead to control issues due to the varying directions of the correction axes caused by the shape and bends in the generated canal. Additionally, the smoothness of the generated canals in GeoSACS is insufficient, especially for tasks like painting, which required special attention.

\section{Methodology}
The goal of our approach is to make control of the robot more intuitive. We propose to do it in three ways: (1) make the canals smoother to avoid natural backtracking when disks are crossing, (2) increase the manipulability where users need it, and (3) provide a more intuitive mapping between user inputs and joystick motions. Our code is available online\footnote{\url{https://gitlab.idiap.ch/hrai/geosacs.git}}.

As robots are expected to interact in carpentered worlds (i.e., environments that have been engineered to have typically flat surfaces and right angles), we assume that users mostly need to interact with horizontal (e.g., table) and vertical (e.g., wall) planes. As such, our method tries to increase manipulability in such areas by ensuring that if the final canal sections are close to a vertical or horizontal plane, they become aligned. Finally, we initially assume that if a user pushes the joystick right, the robot should move toward the ``right'' and if the joystick is pushed forward, the robot should either go forward (if the disk on the canal is near horizontal) or up (if the disk is near vertical).

\subsection{Initial mapping}

\begin{figure}[h]
\centering
\begin{subfigure}{.15\textwidth}
    \centering
    \includegraphics[width=.95\linewidth]{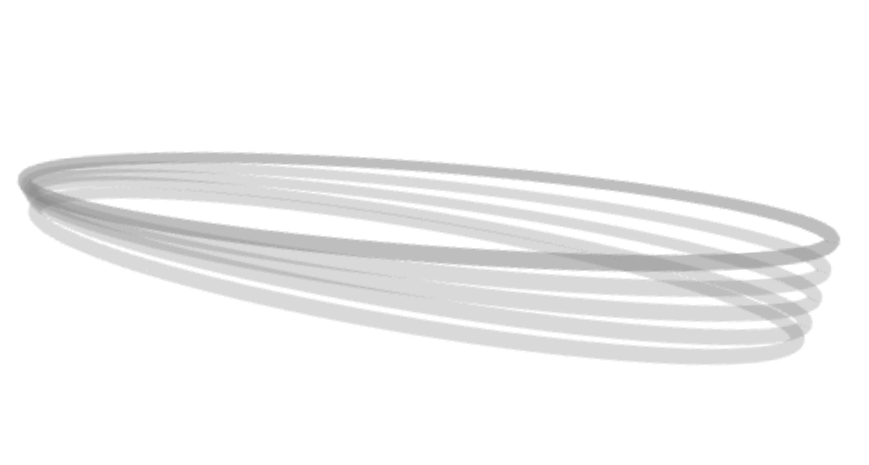}  
    \caption{Near-horizontal disks}
    \label{canal1}
\end{subfigure}
\begin{subfigure}{.16\textwidth}
    \centering
    \includegraphics[width=.7\linewidth]{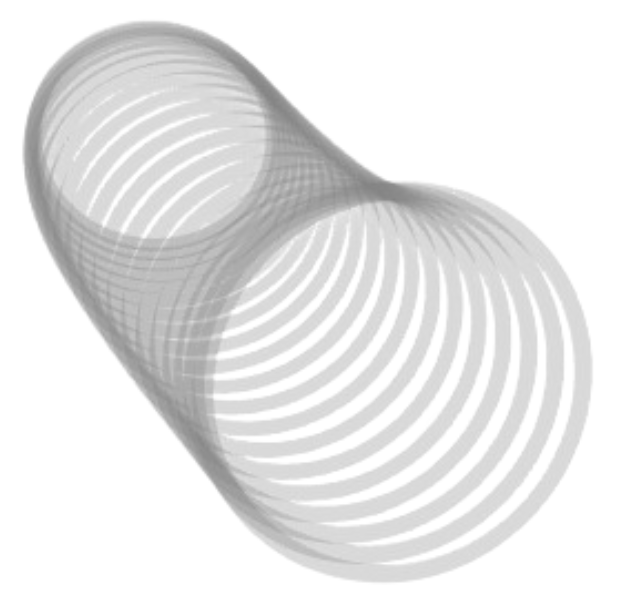}  
    \caption{ \hspace{-2pt} Near-vertical disks facing the user}
    \label{canal2}
\end{subfigure}
\begin{subfigure}{.16
\textwidth}
    \centering
    \includegraphics[width=.85\linewidth]{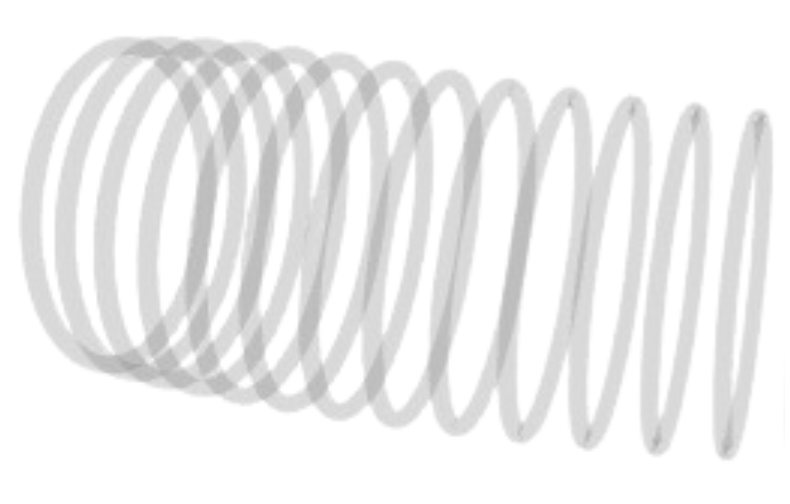}  
    \caption{\hspace{-3pt} Near-vertical disks facing sideways}
    \label{vertical_typeB}
\end{subfigure}

\caption{Categories of disks found within a canal}
\label{fig:disk_types}

\end{figure}

To align user inputs to corrections on a disk, we identify two situations, near-horizontal disks and near vertical disks (see Figure \ref{fig:disk_types}).
When a user issues a directional command via the joystick, the system identifies the relevant disk and extracts the corresponding correction axes (i.e., the correction x-axis $\textbf{x}_{s}$ and y-axis $\textbf{y}_{s}$). For near-horizontal disks, these correction axes are projected onto the ground plane, and their alignments with the joystick axes are calculated using the dot product values and their associated signs. For near-vertical disks, one correction axis is typically more vertical, aligned with the global z-axis, $\textbf{z}_{G}$, while the other is typically more horizontal, aligned with the ground plane. $\textbf{j}_{y}$ is aligned with the ``vertical'' correction axis, so when the user pushes the joystick forward (positive $\textbf{j}_{y}$ direction), the robot moves upward, and vice versa. The remaining correction axis is aligned with $\textbf{j}_{x}$ by projecting onto the ground plane and taking the dot product with $\textbf{j}_{x}$.

\subsection{Pilot Study} \label{pilot_study_proc}

To experiment with and gather feedback on our mapping method, we conducted a pilot study with three participants including three types of tasks: an object relocation task, a laundry loading task, and a painting task (see Figure \ref{fig:data_collection}). The first two participants used only a single condition (initial version of our approach) assessing and refining the method for intuitiveness. We learned that for the near-vertical planes facing sideways (see Figure \ref{vertical_typeB}), users expected movement in one direction when the disk was on their left and the opposite when on their right, with direction remaining consistent on each side and only changing when switching sides. The third participant tested two conditions: the refined mapping framework based on feedback from the first two participants, and the baseline condition, GeoSACS, allowing us to confirm the feasibility of a full study for the formal evaluation.

The canals for the three task types were generated using the pre-processing steps outlined in GeoSACS, with only two demonstrations. 
For the studies, we used the Lio robot, designed specifically for eldercare and home environments by F\&P Robotics\cite{lio_overview}, along with an Xbox gaming controller to capture user inputs.

\subsection{Dynamic Input Mapping Framework}

Building on insights from the pilot studies, we finalized our input mapping strategy. The intuition is that for horizontal disks, the ``right'' and ``forward'' joystick directions are aligned with the robot motions; for vertical disks facing the user, ``forward'' is mapped to ``up'' for the robot, ``right'' is aligned; and for sideways disks, ``forward'' is mapped to ``up'', and ``right'' will be mapped to ``forward'' if the disk is on the user's left side, and to ``back'' if the disk is on the right side. For a more precise explanation, refer to Algorithm \ref{alg:input_mapping} with the following notations: the disk where the correction is applied denoted as \(C_{s}\), the correction axes on \(C_{s}\) as \(\textbf{x}_{s}\) and \(\textbf{y}_{s}\), the coordinates of the directrix point at the center of \(C_{s}\) in global frame as \(d_{s}\), and the tangent vector to the directrix at that point as \(\textbf{e}_{T}(s)\). Finally, the aligned correction axis and direction for \(\textbf{j}_{x}\) are denoted as \(\textbf{A}_{x}\) and \(D_{x}\), and for \(\textbf{j}_{y}\), as \(\textbf{A}_{y}\) and \(D_{y}\), with a projection function indicating the projection to the ground.

\begin{algorithm}[t]
\caption{Dynamic Input Mapping}\label{alg:input_mapping}
\begin{algorithmic}[1]
    \State \textbf{Input:} $\textbf{x}_s, \textbf{y}_s, \textbf{j}_x, \textbf{j}_y, \textbf{e}_T(s), \textbf{z}_G, d_s$
    \State \textbf{Output:} $\textbf{A}_x, \textbf{A}_y, D_x, D_y$  

    \State $ \theta \gets \arccos{ \left( \frac{\textbf{e}_T(s) \cdot \textbf{z}_G}{|\textbf{e}_T(s)| |\textbf{z}_G|} \right) }$
    \If{$ \frac{\pi}{3} <$  $\theta < \frac{2\pi}{3}$}
        \State $disk\_orientation \gets ``\textbf{near-horizontal}"$
    \Else
        \State $disk\_orientation \gets ``\textbf{near-vertical}"$
    \EndIf

    \If{$disk\_orientation$ is ``\textbf{near-horizontal}" }
        \State $\textbf{p}_x, \textbf{p}_y \gets \Call{projection}{\textbf{x}_s, \textbf{y}_s}$

        \State $\textbf{A}_x, D_x \gets (|\textbf{j}_x \cdot \textbf{p}_x| > |\textbf{j}_x \cdot \textbf{p}_y|) \,?\, (\textbf{x}_s, \Call{Sign}{\textbf{j}_x \cdot \textbf{p}_x}) \,:\, (\textbf{y}_s, \Call{Sign}{\textbf{j}_x \cdot \textbf{p}_y})$

        \State $\textbf{A}_y, D_y \gets (|\textbf{j}_y \cdot \textbf{p}_x| > |\textbf{j}_y \cdot \textbf{p}_y|) \,?\, (\textbf{x}_s, \Call{Sign}{\textbf{j}_y \cdot \textbf{p}_x}) \,:\, (\textbf{y}_s, \Call{Sign}{\textbf{j}_y \cdot \textbf{p}_y})$

    \ElsIf{$disk\_orientation$ is ``\textbf{near-vertical}" }


        \If{$| \textbf{y}_s \cdot \textbf{z}_G | > | \textbf{x}_s \cdot \textbf{z}_G |$}
            \State $\textbf{A}_y \gets \textbf{y}_s$, $\textbf{A}_x \gets \textbf{x}_s$, $D_y \gets \Call{Sign}{\textbf{y}_s \cdot \textbf{z}_G}$
        \Else
            \State $\textbf{A}_y \gets \textbf{x}_s$, $\textbf{A}_x \gets \textbf{y}_s$, $D_y \gets \Call{Sign}{\textbf{x}_s \cdot \textbf{z}_G}$
        \EndIf

        \State $\textbf{p}_x \gets \Call{projection}{\textbf{A}_x}$
       
        \State $\theta_{\textbf{A}_x} \gets \arccos{ \left( \frac{\textbf{p}_x \cdot \textbf{j}_y}{|\textbf{p}_x| |\textbf{j}_y|} \right) }$

        \If{$ \frac{5\pi}{6} > \theta_{\textbf{A}_x} > \frac{\pi}{6}$} \Comment{The disk is facing the user}
            \State $D_x \gets \Call{sign}{\textbf{j}_x \cdot \textbf{p}_x}$
        \Else \Comment{The disk is facing sideways}
        

\State $P_x, P_y, P_z \gets \textbf{R}^\top \cdot d_s, \, \textbf{R} \gets [\textbf{j}_x \, \textbf{j}_y \, (\textbf{j}_x \times \textbf{j}_y)]$

            \State $D_x \gets (P_x < 0) \,? \, \Call{sign}{\textbf{j}_y \cdot \textbf{p}_x} \,:\, -\Call{sign}{\textbf{j}_y \cdot \textbf{p}_x}$
        \EndIf

    \EndIf

    \State \Return $\textbf{A}_x, \textbf{A}_y, D_x, D_y$

\end{algorithmic}
\end{algorithm}

\subsection{Other Improvements}
\label{sec:other}

\subsubsection{Smoothing Canal Surfaces}

To simplify manipulations on horizontal and vertical planes at the extremities of the canal for pick-and-place or painting, 
we ensured that the tangent vectors near the ends would be orthogonal to either a horizontal or vertical plane, and we perform an optimization to smoothen the transition between the aligned disks and the non-aligned ones. We start by classifying the canal end 
\(E_s\) as either vertical or horizontal, by computing \(\theta\), the average angular distance between the mean direction of the last 10 tangent vectors and the global z axis \(\textbf{z}_{G}\) and then align it with the closest axis.

Based on this classification, we force the last part of the canal (set empirically to 20\% in our experiments) to be either vertical or horizontal.
This guarantees that the robot moves through these regions along clean vertical or horizontal paths, simplifying user control and interpretability. 
Due to this forced alignment of the disks at the ends, intersections between adjacent disks may occur, that we minimize through optimization.

\subsubsection{Moving Out of Canals}

During the pilot studies, participants also mentioned that it would be helpful if the robot could move slightly beyond its current range when needed. This issue was especially noticeable during the laundry task, where clothes would sometimes hang from the edge of the laundry drum. When participants attempted to move the robot down to push the clothes further inside, it did not respond as they expected. This limitation occurred because, during the demonstrations, it was difficult to demonstrate an exact margin without risking a collision. To address this, we implemented a method that allows users to move partially outside the generated canal. However, we limited this extension to prevent any safety issues that could arise from unrestricted movement beyond the canal. For safety purposes, after the correction is applied, we gradually shrink the radius, ensuring that after a window of 10 disks (which in practice is less than 5cm), the robot returns to moving along the canal's boundary.

\section{Comparative User Study}
Our first goal in this article is to evaluate the new mapping system proposed here. To do so, we conducted a user study with 20 abled participants recruited from the research institution, and mixing both technical and administrative staff. The study aims to compare our method to a GeoSACS baseline. However, it is important to note that since GeoSACS lacks a method to refine the canal and ensure the disks are vertical as in our case, for the painting task, we had to manually adjust the disk near the painting board to be perfectly vertical.

\subsection{Tasks}

Similarly to our pilot study, we used three different tasks to evaluate our method. Participants had up to eight minutes to complete each task, or could stop when they considered having finished the task.

\subsubsection{Object Relocation Task}: In this task (see Figure \ref{fig:data_collection} top row), participants are asked to move three objects randomly located from a table on one side of the robot to a location marked on a printed sheet on a second table on the other side of the robot. The motivation for this task is to evaluate the performance in general tabletop pick-and-place tasks that could be encountered in daily life. Two demonstrations were conducted in a way that the canal is generated to link the two tables, allowing users to guide the robot in placing the objects at their intended goal locations.

\subsubsection{Painting Task}: In this task (see Figure \ref{fig:data_collection} middle row), participants are asked to pick up brushes with custom-designed holders on a table (pink, blue, and yellow) and use them to paint as they wish on a paper sheet on the wall in front of the robot. This task is inspired from \cite{sheidlower2024online} supporting the idea that assistive robots should also support creativity. As such, this task's evaluation is primarily subjective. Due to the challenge of picking up brushes, participants were instructed to grasp the paint brush as best as they could, and the experimenter would manually check the brush and ensure it is properly grasped, allowing participants to focus on their painting. To switch brushes, participants could drop them on a plate before picking another one. If needed, experimenters would refill the container and replace the brush inside, restoring it to the original position. The canal was created from two demonstrations to cover the three brush holders and the plate on one side, and the vertical paper sheet on the other side. 

\subsubsection{Laundry Loading Task}
\begin{figure}[t]
    \centering
    \includegraphics[scale=0.175]{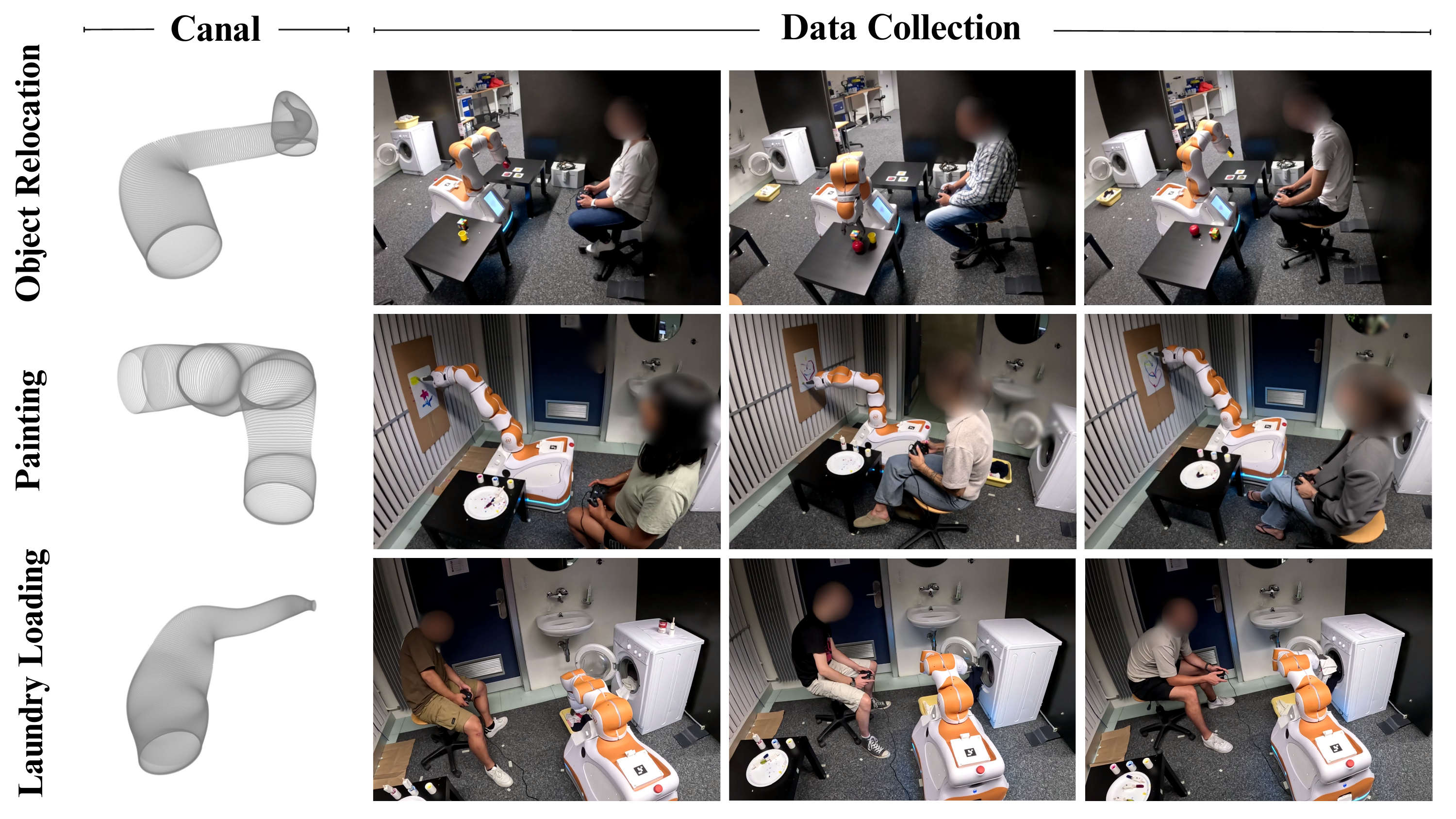}
    \caption{Data collection from participants including the generated canals for the three tasks.}
  \label{fig:data_collection}
  
\end{figure}

: In the last task, participants were instructed to load five clothing pieces from a basket to a laundry machine (see Figure \ref{fig:data_collection} bottom row). The goal of this task is to evaluate the system's performance in real ADLs with 3D interaction. The positions of the clothing items were randomized between experiments. Participants were instructed to push any clothes hanging out of the laundry drum door after their initial attempt to place them inside. The canal was created from two demonstrations to cover the laundry basket on the ground and to go inside the laundry machine.

\subsection{Procedure}

The study procedure was reviewed and approved by the ethics committee of our institution. We employed a within-subjects design, with the two methods counterbalanced to minimize order effects, and the task order was fixed (object relocation, painting, and laundry loading).

At the beginning, participants were given a brief introduction to the study and the tasks involved, followed by obtaining their informed consent. Participants then completed a demographic questionnaire, which included questions on a 5-point Likert scale about their prior experience with robots and gamepads, such as an Xbox joystick. All participants were compensated CHF 37.5 for their participation in the study, and the study lasted around one and half hours.

We then moved into the training phase, with the goal of familiarizing participants with the gamepad controls. Our setup consisted of four joystick buttons: one to start, one to change the movement direction within the canal, one to open/close the gripper, and one to stop. Additionally, the left 2D joystick of the Xbox gamepad was used for providing corrections on the disks. The controls remained the same for both methods, with the only difference being the robot's movements resulting from the online mapping mechanism. The training was structured as a tabletop pick-and-place task, where we placed a can on one table and a paint brush with the custom-designed holder in an empty container, and introduced the control mechanism to the participants. During the training phase, the object location markers for the object relocation task were covered with paper. Participants were given 8 minutes to freely explore the controls and observe their effect on the robot. Most participants picked up the can and placed it on the other table, while some attempted to grasp the paint brush from the holder and place it in the can on the other table. Afterward, they proceeded to complete the tasks in the order previously mentioned. After finishing a set of tasks, participants answered a questionnaire (see next section). This process, including training, tasks, and questionnaire, was followed for the second condition. Finally, the experiment concluded with a semi-structured interview, debriefing, and compensation.

\subsection{Variables}

\subsubsection{Independent Variables} Our study has two conditions: (1) the GeoSACS SA baseline (with the painting canal verticalized), and (2) our approach with additional input mapping and canal smoothing.%

\subsubsection{Dependent Measures - Objective} 
For both the object relocation and laundry loading tasks, we measured the time taken to complete the task and the proportion of time spent on providing corrections relative to the total task time. For the object relocation task, our performance metric was the number of objects correctly positioned at the target location within the allotted time. For the laundry loading task, we implemented a point-based scoring system: participants received two points for each piece of clothing fully placed inside the laundry drum, one point if part of the clothing was hanging from the laundry drum door, and zero points if the clothing remained in the basket or fell on the floor where it could no longer be picked up. Thus, the maximum score a participant could achieve for the five clothing pieces was 10 points.

\subsubsection{Dependent Measures - Subjective} 
At the end of each condition, we asked participants to complete a 7-point Likert scale using the USE questionnaire \cite{use}, focusing on the sections for ease of use, ease of learning, and satisfaction, for a total of 22 questions. Additionally, participants completed the NASA-TLX questionnaire \cite{TLX} to assess the workload associated with each method. A semi-structured interview was conducted at the very end of the study to gather feedback on usability, efficiency, perceived safety, workload, and overall impressions of the method.

\subsection{Hypothesis}
Our evaluation consisted of the following two hypotheses (with their associated predictions):%

\textbf{H1:} Users will complete tasks efficiently using GeoSACS combined with the mapping framework compared to using GeoSACS alone.

\begin{enumerate}
    \item[] \textbf{P1.1:} Users will complete object relocation task in less time with our method.
    \item[] \textbf{P1.2:} Users will complete object relocation task with fewer corrections using our method.

    \item[]  \textbf{P1.3:} Users will position objects more accurately in the object relocation task using  our method.

    \item[] \textbf{P1.4:} Users will achieve higher scores for placing clothes inside the laundry machine with our method.
\end{enumerate}

As we anticipated that not all participants would complete the laundry loading task within the allocated time, we excluded times related to the laundry task from our predictions.

\textbf{H2:} GeoSACS combined with the mapping framework will be more intuitive than GeoSACS alone.
\begin{enumerate}
    \item[] \textbf{P2.1:} Users will report higher usability with our method.
    \item[] \textbf{P2.2:} Users will report lower workload with our method.
\end{enumerate}

\subsection{Population}

The comparative study was conducted with 20 abled participants (twelve male, eight female) aged 25 to 48 (Mean = 30.65, SD = 5.9268), primarily recruited from our research institute, with one participant recruited externally by word of mouth. Our participants were composed of a mixture of technical staff and administrative staff. The mean values for prior experience with robots and gamepads were observed to be 2.05 $\pm$ 1.36 and 3.30 $\pm $ 0.95, respectively, based on responses collected using a 5-point Likert scale, where lower scores indicate less prior experience.

\subsection{Results}

The results for the object relocation and laundry loading tasks are presented in Table \ref{results}. We used a Wilcoxon signed-rank test to calculate the p-values, as a Shapiro-Wilk test indicated that the data did not follow a normal distribution, except for the correction time in the laundry loading task, where the data was normally distributed, and where we conducted a paired t-test. These results supported \textbf{P1.1} but did not support \textbf{P2.2}, despite a trend toward significance. A possible explanation, observed during the studies, is that some participants frequently issued corrections in both conditions instead of utilizing the robot's autonomous routines effectively.

Moreover, all 20 participants successfully placed all three objects using our method, compared to an average score of 2.55 (SD = 0.58) for GeoSACS alone, with a Wilcoxon test p-value below 0.01 (Z = 0), supporting \textbf{P1.3}. The discrepancy in performance stemmed from the trial-and-error approach used with GeoSACS, where, during object picking, the end effector would occasionally disturb nearby objects. This made it harder to grasp these objects later, as some would roll into positions where they could no longer be retrieved.

For the laundry loading task, our method achieved a higher average score of 9.26 (SD = 0.96) compared to 8.26 (SD = 0.95) for GeoSACS (Z = 2.262, p $<$ 0.01 (Wilcoxon)), supporting \textbf{P1.4}. One participant's data was excluded from the analysis, as they spent excessive time pushing clothes into the machine rather than following the instructions to first place as many clothes as possible. This participant was identified as an outlier using Tukey's fences, which justified their removal from the final calculation. Together, these results provide partial support for hypothesis \textbf{H1}.

\begin{table}[]
\caption{Completion and correction times for the two methods in the object relocation and laundry loading tasks.}
\label{results}
\begin{tabularx}{\columnwidth}{p{1.0cm} p{1.9cm} X X X X X}
\toprule
\multicolumn{1}{l}{\multirow{2}{*}{Task}} & \multicolumn{1}{l}{\multirow{2}{*}{Metric}} & \multicolumn{2}{l}{GeoSACS} & \multicolumn{2}{l}{Ours} & \multirow{2}{*}{p-Val} \\ \cmidrule(lr){3-6}
\multicolumn{1}{l}{} & \multicolumn{1}{l}{} & \multicolumn{1}{l}{Mean} & \multicolumn{1}{l}{SD} & \multicolumn{1}{l}{Mean} & \multicolumn{1}{l}{SD} &  \\ 
\midrule
\multirow{2}{*}{\makecell[l]{Object \\ relocation}} & Completion time & 259.02 & 71.05 & 220.45 & 36.60 & $<$0.05 \\
& Correction time & 69.29 & 52.94 & 54.70 & 30.22 & 0.058 \\
\midrule
\multirow{2}{*}{\makecell[l]{Laundry \\ loading}} & Completion time & 427.13 & 69.70 & 385.84 & 75.50 & 0.053 \\
& Correction time & 121.43 & 44.05 & 107.57 & 42.64 & 0.26 \\
\bottomrule
\end{tabularx}
\end{table}

\begin{figure*}[t]
    \centering
    \begin{subfigure}[b]{0.72\textwidth}
        \centering
        \includegraphics[scale=0.25]{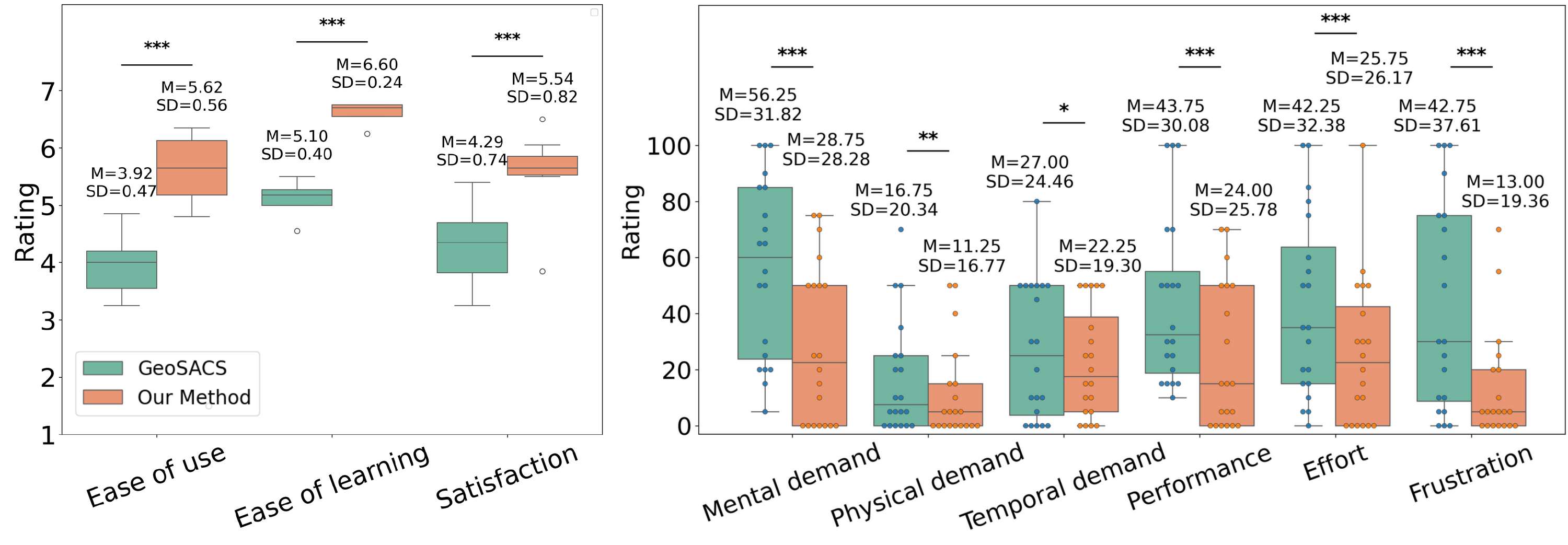}
        \caption{Participant ratings for ease of use, ease of learning, and overall satisfaction (higher the better with p-values calculated using paired t-test). Workload was assessed across the six dimensions of the NASA-TLX scale (lower the better with p-values calculated using Wilcoxon test).  [(\textbf{*}) denotes p $<$ 0.05, (\textbf{**}) denotes p $<$ 0.01, and (\textbf{***}) denotes p $<$ 0.001]. }
        \label{fig:subjective}
    \end{subfigure}%
    \hfill
    \begin{subfigure}[b]{0.25\textwidth}
        \centering
        \vspace*{\fill}
        \raisebox{.20\height}{\includegraphics[height=1.45in]{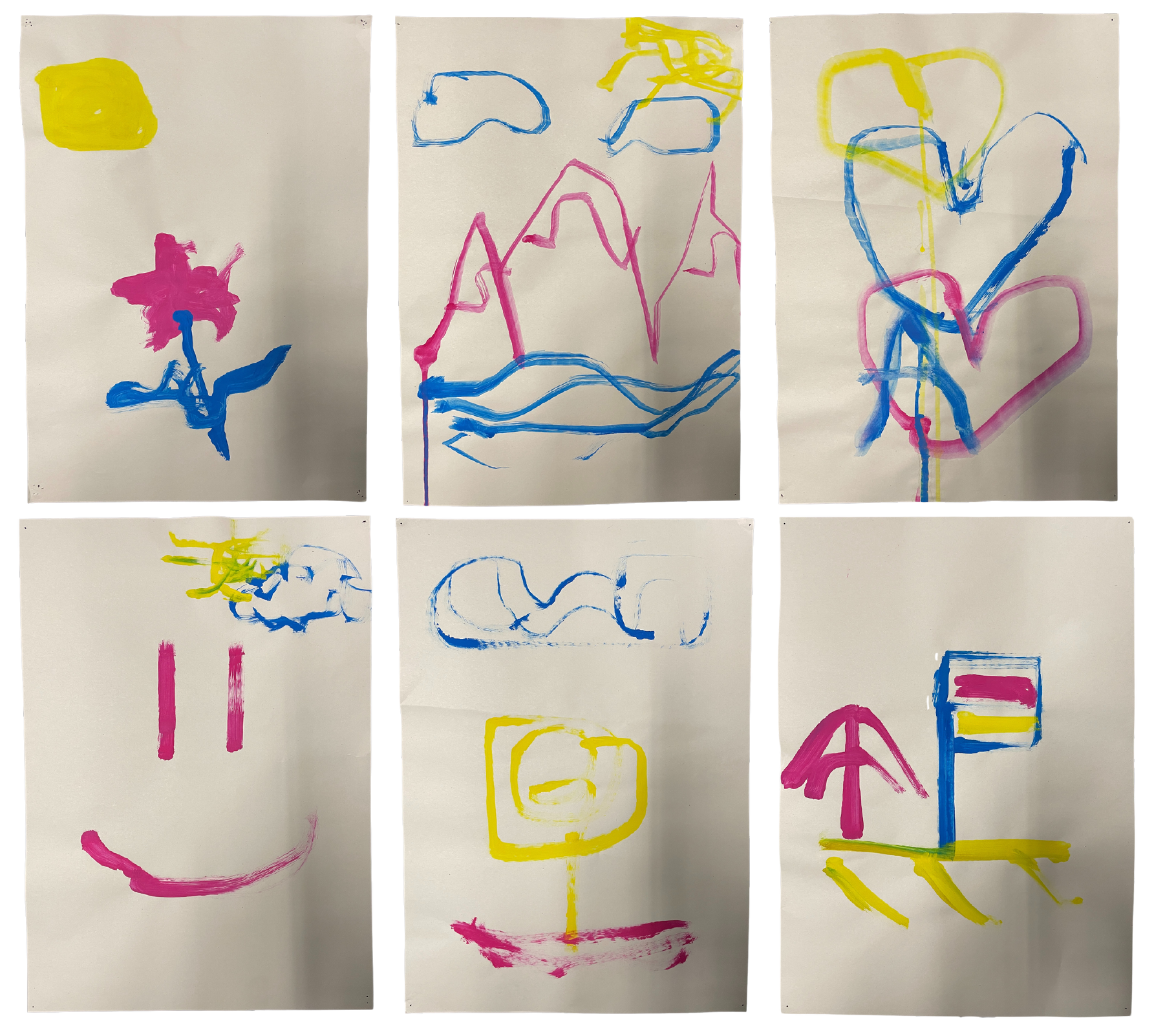}}
        \vspace*{\fill}
         \caption{Some meaningful paintings done by participants using our method within 8 minutes. They successfully used all three colors.}
  \label{fig:paintings}
    \end{subfigure}

    \caption{Subjective results from the comparative study conducted with 20 participants.}
    \label{fig:experiments}
\end{figure*}

Regarding subjective measures, for the painting task, all participants preferred our method. Except for one, all participants used all three colors to complete their paintings. With GeoSACS, participants generally struggled to create anything meaningful. In contrast, with our method, several participants were able to produce recognizable paintings, such as flowers with a sun in the background, hearts, smiley faces, and even letters (see Figure \ref{fig:paintings}).
Figure  \ref{fig:subjective} shows participants' evaluations of ease of use, ease of learning, and satisfaction for both methods, as well as the ratings across the six dimensions of the NASA-TLX. The results indicate that our method consistently outperforms GeoSACS across all measured metrics, validating our predictions \textbf{P2.1} and \textbf{P2.2}, and ultimately supporting \textbf{H2}.

The results from the semi-structured interviews conducted at the end of the experiments further reinforced our findings. Every participant preferred our method for all tasks, except for one participant who favored GeoSACS for the laundry loading task. For the intuitiveness of the mapping, several participants remarked, ``The second method [ours] was more intuitive", ``I liked the first method [ours] very much", ``It is intuitive to use", ``It was great", and ``The first one [ours] felt natural way to control", further confirming the intuitiveness of our mapping framework. Regarding the ease of use of our method, a few participants commented, ``It was easy to use", "You don't even have to think", and ``The first one [ours] was quite easy".

Several participants specifically commented, ``This is like playing a game", ``It was easy to use, friendly, and fun", and ``It was really fun to use". These sentiments, coupled with 12 participants selecting ``strongly agree" for the question ``It is fun to use" in the satisfaction section of the USE questionnaire, highlight that our method is not only effective but also enjoyable for users.

Regarding suggestions, most participants recommended adding a pause button to stop the robot at any point and then resume it by pressing the same button again. Another common suggestion was to provide control over the robot’s execution speed. Several participants proposed the option to slow down in areas of high uncertainty, such as near tables, the laundry basket, or the painting board, which would allow for more precise task completion. Lastly, a few participants suggested introducing a separate button to rotate the end effector, enabling them to adjust the grip as needed, rather than being limited to the grasp orientation demonstrated initially.

\section{Accessibility Work}

We then explored how to adapt our method for users with disabilities, the population for which assistive robots would make most sense.
For example, we previously used a standard Xbox gamepad in the comparative study, which may not be well-suited for individuals with disabilities due to its small buttons and joysticks limiting accessibility. We first looked for wheelchair inspired joysticks, but mostly found limited, costly, and older devices. We then explored the work done in the accessible gaming community \cite{grammenos2009designing}, which has explored more accessible interfaces for disabled gamers \cite{maggiorini2019evolution}, and finally found the recently released Sony Access Controller\footnote{Sony Access Controller: \url{https://www.playstation.com/en-gb/accessories/access-controller/}}. This controller is a specialized gamepad with customizable buttons and a larger and easier to manipulate joystick closer to those found in electric wheelchairs, and as such, we decided to use it for our future work.

\section{Exploratory study with wheelchair users}
Based on the results of our comparative study and accessibility work, we conducted an exploratory study of our approach with wheelchair users to evaluate it with a population closer to our target audience. This served as an initial evaluation before implementing further design improvements in future, which will be guided by a participatory design approach aimed at bringing SA-based assistive robotics to real users who could benefit from it.

\subsection{Population}

This study was conducted with 3 participants, all wheelchair users, two used manual wheelchairs with one using additional electric assistance, and one used an electric wheelchair but could temporarily walk. One participant also brought their partner to the study site. For local ethical reasons, we cannot report health data of participants. We recruited participants from the local community by distributing flyers to local groups and therapists. Participants were aged 52 to 61 (Mean = 56.67, SD = 3.68) with two females and one male. The study lasted around one hour and half and was compensated with CHF 50.

\subsection{Procedure}
The procedure followed was similar to the comparative study, with the same consent process, training, and tasks in the same order. However, we made the following key changes for accessibility: (1) we only evaluated our method, (2) we did not impose any time limit, (3) the experimenter could provide some occasional verbal guidelines and encouragements, and (4) we did not ask them to complete the full questionnaires as it can be tiring and complicated to do, instead we used these questions to drive our semi-structured interview.

\subsection{Results and Observations}
All the wheelchair participants were successful in the relocation task, however for two participants, one object needed to be replaced in their initial position after being pushed accidentally. The painting task proved a bit more challenging, participants tended to give themselves short-term objectives (e.g., drawing a spiral or filling a rectangle) and achieved partial success. For the laundry tasks, participants scored 9, 9, and 10 without the experimenter help (but the robot had to be restarted once due to a collision and network issue).

During the study, one participant commented that the gamepad reminded them of a wheelchair joystick, which was the intention behind the selection. Another participant was talking to the robot, providing it with encouragement and instructions. However, a participant also reported challenges to see perspectives, which impacted their use of the robot. During the debrief session with the experimenter, they suggested that having access to the front camera could be useful.

One participant stated that such a system could already be useful, especially for laundry, as both loading the machine and removing clothes from a drying rack are challenging. Others mentioned that it could be useful to lift heavy objects from the ground or access things high up. However, all participants reported that it would be more useful for people with severe mobility limitations: ``with this type of things we could do a lot of things, especially for disabled people with [heavier pathologies]", another participant even mentioned that they would recommend the study to a friend. %

Two participants mentioned that the system was easy to use, and while painting was reported as being more challenging, they reported they enjoyed it, and one mentioning that it would take a few trials to get better at it. The last participant had a few more challenges, partially due to the perspective and a perceived high temporal load, when actions needed to be synchronized: ``when it worked for the first time, I was very happy, [...] but when it didn't work, [I was] more lost''. They compared their experience with electric wheelchair, stating that it would be useful to adapt the speed to have fine control when needed. Additionally, they suggested that for home deployments, it would be useful to have a cheat sheet with the different buttons.

An interesting point is that participants assigned agency to the robot, talking to it, using expressions like needing to ``give it some time", ``it did what it wanted'', or ``[when I was rushing it] it felt like it was losing its memory''. 
 Overall, every participant mentioned it was fun and volunteered to participate in more studies.

\section{Discussion}\label{AA}

The results from the comparative study indicate that our method holds significant promise for handling complex tasks. Additionally, the paintings created by participants demonstrate the feasibility of our approach for creative and intricate tasks. Moreover, the initial exploratory study with wheelchair users highlighted strong engagement and the potential of our method for individuals with disabilities. However, further refinements through participatory design and more rigorous evaluations are necessary to fully realize its potential.

\subsection{Insights}

Overall, participants found the method to be flexible and intuitive to control. During the study, they appeared relaxed due to the robot's autonomous behavior, allowing them to intervene only when necessary. Many participants even engaged in conversation with the robot, offering comments such as ``Bravo Lio”, ``No no no”, and ``Oh, sorry Lio” which suggests that a more interactive environment could be developed with feedback mechanisms, including verbal responses from the robot.
Interestingly, a few participants initially expressed apprehension when introduced to the tasks. However, by the end of the study, all participants reported feeling confident about the robot's safety and their own.
This feedback indicates that our SA approach not only enabled convenient robot control, but also increased user confidence to use such systems.

Our exploratory study with wheelchair users provided initial support toward the usability of our system by this population, and provided us additional knowledge about the potential users benefiting most from the system as well as design improvements.

\subsection{Limitations and Future Work}

\subsubsection{Limitation of the approach} 
Despite the proposed method allowing some flexibility to move outside the generated canal, the robot's motions are still largely confined to the canal structure. This limitation was particularly noticeable when objects or clothes accidentally fell into areas outside the robot’s reach due to the canal's constraints. Additionally, participants sometimes struggled to maintain a clear view of the task, with their line of sight obstructed by the robot’s body. This issue was especially evident during the laundry and painting tasks, where participants were seen shifting their heads to gain a better perspective, which was especially challenging for wheelchair users. Moreover, further refinements are necessary for the system to handle tasks requiring precise manipulation. The current system also lacks environmental awareness and still relies on kinesthetic demonstrations. 

For future work, we plan to integrate vision-based techniques to enable the generation of demonstration-free, lightly constrained canals, and communicate more information to the user, e.g., by displaying the robot view. Building on participant feedback, we also aim to develop a velocity adjustment mechanism and an end effector orientation control system, which will enhance the robot's capabilities for more precise manipulation tasks.

\subsubsection{Limitation of the study}
Our study also contained limitations. First, it only compared our approach to a single baseline and in a short-term interaction. We made this decision as other SA frameworks can be challenging to adapt to complex 3D tasks as the one explored in this paper. Furthermore, the comparative study was limited to 20 participants, and we only recruited 3 wheelchair users in our exploratory study. To enhance the validity of our findings, we plan to involve a broader and more diverse population in our future work.

Additionally, we plan to conduct more in depth participatory design with wheelchair users in the future to improve the accessibility, usability, and usefulness of our approach for this population. Finally, the willingness of our wheelchair participants to engage in future research activities is a positive sign for the next steps of this research. 

\section{Conclusion}\label{AA}

In this paper, we presented the design and evaluation of our input mapping mechanism for controlling high-DoF robots, aimed at enhancing usability of assistive robots. A comparative user study with 20 abled participants across a range of tasks demonstrated the efficiency, usability, and reduced workload enabled by our method compared to a baseline. Additionally, we conducted an exploratory study with 3 wheelchair users, utilizing a specialized gamepad with larger buttons and joystick similar to those found on wheelchairs. The results from this study, serving as a first step toward a participatory design approach, suggest that our method shows great potential for assisting individuals with disabilities.

\bibliographystyle{IEEEtran} 
{\footnotesize
\bibliography{refs} 
}

\end{document}